\documentclass{article}

\usepackage{aaai19}  
\usepackage{times}
\usepackage{helvet}
\usepackage{courier}
\usepackage[utf8]{inputenc} 
\usepackage[T1]{fontenc}    
\usepackage{url}            
\usepackage{booktabs}       
\usepackage{nicefrac}       
\usepackage{microtype}      
\usepackage{amsmath,amssymb,amsfonts,amsthm}
\usepackage[export]{adjustbox}
\usepackage{caption}
\usepackage{subcaption}
\usepackage{algorithm}
\usepackage[noend]{algpseudocode}
\usepackage[usenames,dvipsnames]{xcolor} 
\usepackage{xspace}
\usepackage{booktabs} 
\usepackage{tabulary}
\usepackage{multirow}
\usepackage{csquotes}
\usepackage{colortbl}
\usepackage{nccmath}

\definecolor{lightred}{rgb}{1,0.8,0.8}
\definecolor{lightgreen}{rgb}{0.8,1,0.8}
\newcommand{\cmark}{\cellcolor{lightgreen}\checkmark}%
\newcommand{\xmark}{{\cellcolor{lightred}{$\times$}}}%

\newtheorem{theorem}{Theorem}[section]
\newtheorem{corollary}{Corollary}[section]
\newtheorem{lemma}{Lemma}[section]
\newtheorem{definition}{Definition}[section]
\newtheorem{assumption}{Assumption}[section]
\newtheorem{proposition}{Proposition}[section]
\newtheorem{supplemma}{Supp. Lemma}[section]

\newenvironment{proofsketch}{%
\proof}{\endproof}

\newcommand{\xxnote}[3]{}
\ifx\hidenotes\undefined
  \usepackage{color}
  \renewcommand{\xxnote}[3]{\color{#2}{#1: #3}}
\fi

\newcommand{\norm}[2]{\left|\left| #1 \right|\right|_{#2}}
\newcommand{\abs}[1]{\left|#1 \right|}

\newcommand{\bbm}{\begin{bmatrix}}
\newcommand{\ebm}{\end{bmatrix}}

\newcommand{\pair}[2]{\left( #1, #2\right)}

\newcommand{\setst}[2]{\left\lbrace #1\;\middle|\;#2\right\rbrace}

\newcommand{\ts}[1]{^{\mathrm{#1}}}

\newcommand{\BAMDP}[0]{\mathcal{M}}
\newcommand{\beliefMDP}[0]{\mathcal{B}}

\newcommand{\state}[0]{s}
\newcommand{\stateSpace}[0]{S}

\newcommand{\action}[0]{a}
\newcommand{\actionSpace}[0]{A}

\newcommand{\stateaction}[0]{z}

\newcommand{\latent}[0]{\phi}
\newcommand{\latentSpace}[0]{\Phi}

\newcommand{\stateSpaceAug}[0]{X}
\newcommand{\transitionFn}[0]{\Omega}

\newcommand{\belief}[0]{b}
\newcommand{\beliefSpace}[0]{B}

\newcommand{\estimator}[0]{\tau}

\newcommand{\rewardFn}[0]{R}
\newcommand{\reward}[0]{r}

\newcommand{\estimate}[0]{x}
\newcommand{\onehot}[1]{e_{#1}}

\newcommand{\policy}[0]{\pi}

\newcommand{\hor}[0]{T}

\newcommand{\alg}[0]{\mathcal{A}}
\newcommand{\sample}[0]{m}

\newcommand{\V}[0]{V}
\newcommand{\Q}[0]{Q}
\newcommand{\tQ}[0]{\tilde{Q}}
\newcommand{\tQmax}[0]{\tilde{Q}_{\mathrm{max}}}
\newcommand{\Rmax}[0]{\rewardFn_{\mathrm{max}}}
\newcommand{\Vmax}[0]{\V_{\mathrm{max}}}
\newcommand{\Qmax}[0]{\Q_{\mathrm{max}}}

\newcommand{\cover}[0]{\mathcal{N}_{\beliefMDP}}
\newcommand{\sampleSet}[0]{C}

\newcommand{\LR}[0]{L_R}
\newcommand{\LP}[0]{L_P}
\newcommand{\LQ}[0]{L_Q}
\newcommand{\LtQ}[0]{L_{\tilde{Q}}}
\newcommand{\LQB}[0]{L_{QB}}

\newcommand{\dis}[2]{d\pair{#1}{#2}}
\newcommand{\epsD}[0]{\epsilon}

\newcommand{\knownSet}[0]{K}
\newcommand{\knownBeliefMDP}[0]{\beliefMDP_\knownSet}
\newcommand{\escapeEvent}[0]{E_\knownSet}

\newcommand{\tpolicy}[0]{\tilde{\policy}}

\newcommand{\tolVI}[0]{\beta}
\newcommand{\coverReduced}[0]{\mathcal{N}'_{\beliefMDP}}

\newcommand{\tB}[0]{\tilde{B}}

\newcommand{\algName}[0]{\textsc{Bayes-CPACE}\xspace}

\newcommand{\pacbayes}[0]{\textsc{PAC-Bayes}\xspace}
\newcommand{\pacbelief}[0]{\textsc{PAC-Bayes}\xspace}

\newcommand{\pacbayesmdp}[0]{\textsc{PAC-Bayes-MDP}\xspace}

\newcommand{\algQMDP}[0]{\textsc{QMDP}\xspace}
\newcommand{\algPOMDPLite}[0]{\textsc{POMDP-lite}\xspace}
\newcommand{\algCPACE}[0]{\textsc{C-PACE}\xspace}
\newcommand{\algSARSOP}[0]{\textsc{SARSOP}\xspace}

\newcommand{\algPOMDPLiteShort}[0]{\textsc{P-Lite}\xspace}
\newcommand{\algNameShort}[0]{\textsc{BCPACE}\xspace}

\newcommand{\envTiger}[0]{\texttt{Tiger}}
\newcommand{\envChain}[0]{\texttt{Chain}}

\newcommand{\envLightDark}[0]{\texttt{LDT}}

\newcommand{\supp}[0]{supplementary material\xspace}
\newcommand{\supplemref}[1]{Supp. Lemma~\ref{#1}}

\newcommand{\eref}[1]{(\ref{#1})}
\newcommand{\aref}[1]{Algorithm~\ref{#1}}
\newcommand{\sref}[1]{Section~\ref{#1}}
\newcommand{\figref}[1]{Figure~\ref{#1}}
\newcommand{\tabref}[1]{Table~\ref{#1}}
\newcommand{\assref}[1]{Assumption~\ref{#1}}
\newcommand{\lemref}[1]{Lemma~\ref{#1}}
\newcommand{\thmref}[1]{Theorem~\ref{#1}}
\newcommand{\defref}[1]{Definition~\ref{#1}}
\newcommand{\corref}[1]{Corollary~\ref{#1}}
\newcommand{\lineref}[1]{(Line~\ref{#1})}
\newcommand{\linesref}[2]{(Lines~\ref{#1}--\ref{#2})}
\newcommand{\propref}[1]{Proposition~\ref{#1}}

\title{Bayes-CPACE: PAC Optimal Exploration in Continuous Space Bayes-Adaptive Markov Decision Processes}

\author{
  Gilwoo Lee, Sanjiban Choudhury, Brian Hou, Siddhartha S. Srinivasa
  \thanks{\texttt{\{gilwoo,sanjibac,bhou,siddh\}@cs.uw.edu}} \\
The Paul G. Allen Center for Computer Science \& Engineering \\
University of Washington\\
Seattle, WA 98115, USA
}

\begin{document}
\maketitle



\begin{abstract}
We present the first PAC optimal algorithm for Bayes-Adaptive Markov Decision Processes~(BAMDPs) in continuous state and action spaces, to the best of our knowledge. The BAMDP framework elegantly addresses model uncertainty by incorporating Bayesian belief updates into long-term expected return.
However, computing an exact optimal Bayesian policy is intractable. Our key insight is to compute a near-optimal value function by covering the continuous state-belief-action space with a finite set of representative samples and exploiting the Lipschitz continuity of the value function.
We prove the near-optimality of our algorithm and analyze a number of schemes that boost the algorithm's efficiency. Finally, we empirically validate our approach on a number of discrete and continuous BAMDPs and show that the learned policy has consistently competitive performance against baseline approaches.
\end{abstract}


\section{Introduction}

Addressing uncertainty is critical for robots that interact with the real world.
Often though, with good engineering and experience, we can obtain reasonable regimes for uncertainty, specifically \emph{model uncertainty}, and prepare \emph{offline} for various contingencies. However, we must to predict, refine, and act \emph{online}.
Thus, in this paper we focus on uncertainty over a set of scenarios, which requires the agent to balance exploration (uncertainty reduction) and exploitation (prior knowledge).

We can naturally express this objective as a Bayes-Adaptive Markov Decision Process~\cite{kolter2009near}, which incorporates Bayesian belief updates into long-term expected return.
The BAMDP framework formalizes the notion of uncertainty over multiple latent MDPs.
This has widespread applications in navigation~\cite{guilliardautonomous}, manipulation~\cite{chen2016pomdp}, and shared autonomy~\cite{javdani2015hindsight}.

Although BAMDPs provide an elegant problem formulation for model uncertainty, \emph{Probably Approximately Correct} (henceforth PAC)  algorithms for continuous state and action space BAMDPs have been less explored, limiting possible applications in many robotics problems.
In the discrete domain, there exist some efficient online, PAC optimal approaches~\cite{kolter2009near,chen2016pomdp} and approximate Monte-Carlo algorithms~\cite{guez2012efficient}, but it is not straightforward to extend this line of work to the continuous domain.
State-of-the-art approximation-based approaches for belief space planning in continuous spaces~\cite{sunberg2017continuous,guez2014bayes} do not provide PAC optimality.

In this work, we present the first PAC optimal algorithm for BAMDPs in continuous state and action spaces, to the best of our knowledge.
The key challenge for PAC optimal exploration in continuous BAMDPs is that the same state will not be visited twice, which often renders Monte-Carlo approaches computationally prohibitive, as discussed in ~\cite{sunberg2017continuous}.
However, if the value function satisfies certain smoothness properties, i.e. Lipschitz continuity, we can efficiently ``cover'' the reachable belief space.
In other words, we leverage the following property:
\begin{displayquote}
  A set of representative samples is sufficient to approximate a Lipschitz continuous value function of the reachable continuous state-belief-action space.
\end{displayquote}

\begin{figure*}[!t]
  \centering
  \includegraphics[width=0.24\linewidth]{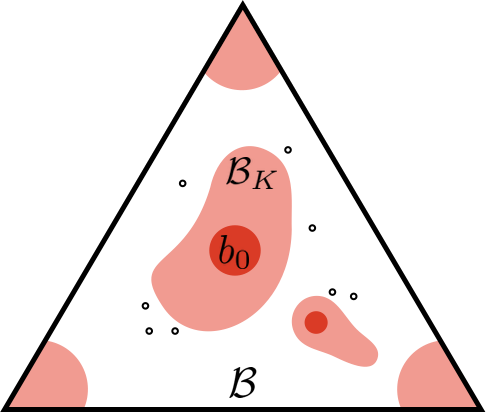} \hfill
  \includegraphics[width=0.24\linewidth]{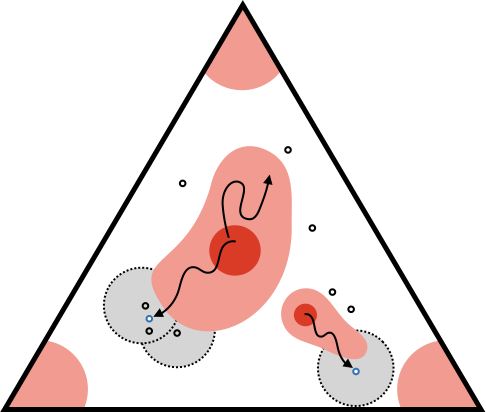} \hfill
  \includegraphics[width=0.24\linewidth]{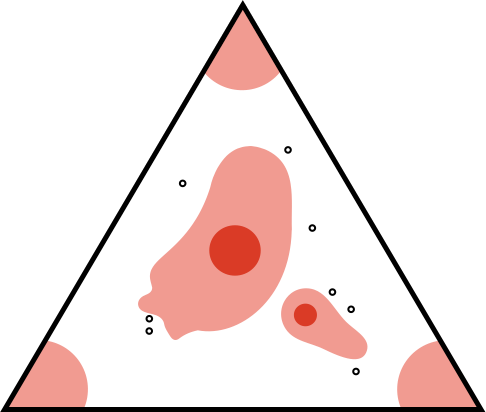} \hfill
  \includegraphics[width=0.24\linewidth]{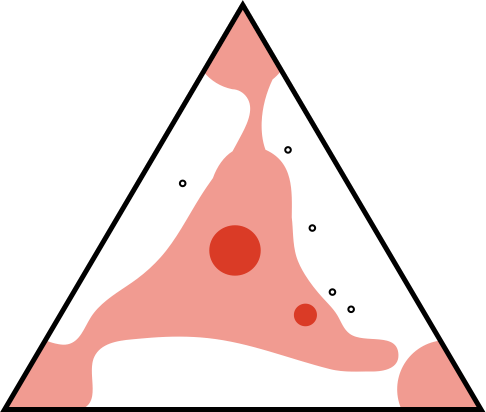}
  \caption{
    The \algName algorithm for BAMDPs.
    The vertices of the belief simplex correspond to the latent MDPs constituting the BAMDP model, for which we can precompute the optimal Q-values.
    During an iteration of \algName, it executes its greedy policy from initial belief $\belief_0$, which either never escapes the known belief MDP $\knownBeliefMDP$ or leads to an unknown sample.
    Adding the unknown sample to the sample set may expand the known set $\knownSet$ and the known belief MDP $\knownBeliefMDP$.
    The algorithm terminates when the optimally reachable belief space is sufficiently covered.
  }
  \label{fig:BCPACE-illustration}
\end{figure*}

Our algorithm, \algName (\figref{fig:BCPACE-illustration}) maintains an approximate value function based on a set of visited samples, with bounded optimism in the approximation from Lipschitz continuity.
At each timestep, it greedily selects an action that maximizes the value function.
If the action lies in an underexplored region of state-belief-action space, the visited sample is added to the set of samples and the value function is updated.
Our algorithm adopts \algCPACE~\cite{pazis2013pac}, a PAC optimal algorithm for continuous MDPs, as our engine for exploring belief space.

We make the following contributions:
\begin{enumerate}
\item We present a PAC optimal algorithm for continuous BAMDPs~(\sref{sec:approach}).
\item
  We show how BAMDPs can leverage the value functions of latent MDPs to reduce the sample complexity of policy search, without sacrificing PAC optimality~(Definitions \ref{def:heuristic_upper_bound} and \ref{def:seeding_exact}).
\item We prove that Lipschitz continuity of latent MDP reward and transition functions is a sufficient condition for Lipschitz continuity of the BAMDP value function~(\lemref{lemma:smooth_action_value}).
\item Through experiments, we show that \algName has competitive performance against state-of-art algorithms in discrete BAMDPs and promising performance in continuous BAMDPs~(\sref{sec:results}).
\end{enumerate}


\section{Preliminaries}

In this section, we review the Bayes-Adaptive Markov Decision Process (BAMDP) framework.
A BAMDP is a belief MDP with hidden latent variables that govern the reward and transition functions.
The task is to compute an optimal policy that maps state and belief over the latent variables to actions.
Since computing an exact optimal policy is intractable~\cite{kurniawati2008sarsop}, we state a more achievable property of an algorithm being \emph{Probably Approximately Correct}. We review related work that addresses this problem, and contrast this objective with other formulations.

\subsection{Bayes-Adaptive Markov Decision Process}\label{ssec:bamdp}

The BAMDP framework assumes that a latent variable $\latent$ governs the reward and transition functions of the underlying Markov Decision Process~\cite{ghavamzadeh2015bayesian,guez2012efficient,chen2016pomdp}.
A BAMDP is defined by a tuple $\BAMDP = \langle \stateSpaceAug, \actionSpace, \transitionFn, P_0, \rewardFn, \gamma \rangle$, where $\stateSpaceAug = \stateSpace \times \latentSpace$ is the set of hyper-states (state $\state \in \stateSpace$, latent variable $\latent \in \latentSpace$), $\actionSpace$ is the set of actions, $\transitionFn(\state', \state, \latent, a) = P(\state' | \state, \latent, a)$ is the transition function, $P_0(\state, \latent)$ is the initial distribution over hyper-states, $R(\state, \latent, \action)$ represents the reward obtained when action $\action$ is taken in hyper-state $(\state,\latent)$, and $\gamma$ is the discount factor.

In this paper, we allow the spaces $\stateSpace, \actionSpace$ to be continuous\footnote{For simplicity of exposition, our notation assumes that the spaces are discrete. For the continuous case, all corresponding probabilities are replaced by probability density functions and all summation operators are replaced by integrals.}, but limit the set of latent variables $\latentSpace$ to be finite. For simplicity, we assume that the latent variable is constant throughout an episode. \footnote{It is straightforward to extend this to a deterministically-changing latent variable or incorporate an observation model. This requires augmenting observation into the state definition and computing belief evolution appropriately. This model is derived in \cite{chen2016pomdp}.}

We now introduce the notion of a \emph{Bayes estimator} $\estimator$.
Since the latent variable $\latent$ is unknown, the agent maintains a belief distribution $\belief \in \beliefSpace$, where $\beliefSpace$ is a $|\latentSpace|$-dimensional probability simplex. The agent uses the Bayes estimator $\belief' = \estimator(\state, \belief, \action, \state')$ to update its current belief $\belief$ upon taking an action $\action$ from state $\state$ and transitioning to a state $\state'$:
\begin{equation*}
  \belief'(\latent) = \frac{\belief(\latent) P(\state' | \state, \latent, \action)}{\sum_{\latent'} \belief(\latent') P(\state' | \state, \latent', \action)}
\end{equation*}

We reformulate BAMDP as a \emph{belief MDP} $\beliefMDP$. We consider the pair $(\state, \belief)$ to be the state of this MDP. The transition function is as follows:
\begin{equation*}
	P(\state', \belief' | \state, \belief, \action) = \left( \sum_{\latent} \belief(\latent) P(\state' | \state, \latent, \action) \right) P(\belief' | \state, \belief, \action, \state')
\end{equation*}
where $P(\belief' | \state, \belief, \action, \state') = 1$ for the belief $\belief'$ computed by the Bayes estimator and zero everywhere else. The reward function is defined as $\rewardFn(\state, \belief, \action) = \sum_{\latent} b(\latent) \rewardFn(\state, \latent, \action)$.

A policy $\policy$ maps the pair $(\state,\belief)$ to an action $\action$. The value of a policy $\policy$ is given by
\begin{equation*}
	\V^\policy(\state, \belief) = \rewardFn(\state, \belief, \action) + \gamma \sum_{\state'} P(\state ' | \state, \belief, \action) \V^\policy(\state', \belief')
\end{equation*}
where $\action = \policy(\state, \belief)$. The optimal Bayesian value function $\V^*(\state, \belief)$ satisfies the Bellman optimality equation
\begin{equation*}
\label{eq:optimal_belief_value}
  \V^*(\state, \belief) = \max_{\action \in \actionSpace} \left[ \rewardFn(\state, \belief, \action) + \gamma \sum_{\state'} P(\state ' | \state, \belief, \action) \V^*(\state', \belief') \right]
\end{equation*}


We now characterize what it means to efficiently explore the reachable continuous state-belief-action space. We extend \cite{kakade2003sample}'s definition of sample complexity for BAMDPs.
\begin{definition}[Sample Complexity]
\label{def:sample_complexity}
Let $\alg$ be a learning algorithm and $\alg_t$ be its policy at timestep $t$. The sample complexity of an algorithm is the number of steps $t$ such that
$\V^{\alg_t}(\state_t, \belief_t) < \V^*(\state_t, \belief_t) - \epsilon$.
\end{definition}

In order to define PAC optimal exploration for continuous space, we need to use the notion of covering number of the reachable belief space.
\begin{definition}[Covering Number]
\label{def:covering_number-prelim}
An $\epsilon$-cover of $\beliefMDP$ is a set $\sampleSet$ of state-belief-action tuples such that for any reachable query $(\state, \belief, \action)$, there exists a sample $(\state', \belief', \action') \in \sampleSet$ such that $\dis{(\state,\belief,\action)}{(\state',\belief',\action')} \leq \epsilon$. We define the covering number $\cover(\epsilon)$ to be the size of the largest minimal $\epsilon$-cover, i.e. the largest $\sampleSet$ which will not remain a cover if any sample is removed.
\end{definition}

Using this definition, we now formalize the notion of PAC optimal exploration for BAMDPs.
\begin{definition}[\pacbayes]
\label{def:pac_bayes}
A BAMDP algorithm $\alg$ is called \pacbayes if, given any $\epsilon > 0$ and $0 < \delta < 1$, its sample complexity is polynomial in the relevant quantities $(\cover(\epsilon), 1/\epsilon, 1/{\delta}, 1/(1-\gamma))$, with probability at least $1-\delta$.
\end{definition}

\subsection{Comparison of \pacbayes vs \pacbayesmdp}\label{ssec:comparison}

We shed some light on the important distinction between the concept of \pacbayes on a BAMDP (which we analyze) and the more commonly referred \pacbayes on an MDP.

The concept of \pacbayes on an MDP with unknown transition and reward functions was first introduced by an online Bayesian exploration algorithm~\cite{kolter2009near}, which is often referred to as BEB~(Bayesian Exploration Bonus) for the reward bonus term it introduces. At timestep $t$, the algorithm forms a BAMDP using the uncertainty over the reward and transition functions of the single MDP being explored at that time. It is assumed that, even when the episode terminates and the problem resets, the same MDP is continued to be explored using the knowledge gathered thus far. The problem addressed is different from ours; \algName produces a policy which is Bayes-optimal with respect to the uncertainty over multiple latent MDPs. We assume that a different latent MDP may be assigned upon reset.

\algPOMDPLite~\cite{chen2016pomdp} extends BEB's concept of \pacbayes to a BAMDP over multiple latent MDPs. Crucially, however, the latent variable in this case \emph{cannot reset} during the learning phase. The authors allude to this as a ``one-shot game ... (which) remains unchanged.'' In other words, \algPOMDPLite is an online algorithm which is near-Bayes-optimal only for the current episode, and it does not translate to a BAMDP where a repeated game occurs.

\subsection{Related Work}

While planning in belief space offers a systematic way to deal with uncertainty~\cite{sondik1978optimal,kaelbling1998planning}, it is very hard to solve in general. For a finite horizon problem, finding the optimal policy over the entire belief space is PSPACE-complete~\cite{papadimitriou1987complexity}. For an infinite horizon problem, the problem is undecidable~\cite{madani1999undecidability}. Intuitively, the intractability comes from the number of states in the belief MDP growing exponentially with $\abs{\latentSpace}$. Point-based algorithms that \emph{sample} the belief space have seen success in \emph{approximately solving} POMDPs~\cite{pineau2003pbvi,smith2005pb}.
Analysis by \citeauthor{hsu2008hardness} shows that the success can be attributed to the ability to ``cover'' the optimally reachable belief space.

Offline BAMDP approaches compute a policy a priori for any reachable state and belief. When $\stateSpace$ is discrete, this is a MOMDP~\cite{ong2010planning}, and can be solved efficiently by representing the augmented belief space with samples and using a point-based solver such as SARSOP~\cite{kurniawati2008sarsop}. A similar approach is used by the BEETLE algorithm~\cite{poupart2006analytic,spaan2005perseus}. \cite{bai2014integrated} presents an offline continuous state and observation POMDP solver which implies it can solve a BAMDP. However, their approach uses a policy graph where nodes are actions, which makes it difficult to extend to continuous actions.

While offline approaches enjoy good performance, they are computationally expensive. Online approaches circumvent this by starting from the current belief and searching forward. The key is to do sparse sampling~\cite{kearns2002sparse} to prevent an exponential tree growth. \cite{wang2005bayesian} apply Thompson sampling. BAMCP~\cite{guez2012efficient} applies Monte-Carlo tree search in belief space~\cite{silver2010monte}. DESPOT~\cite{somani2013despot} improves on this by using lower bounds and determinized sampling techniques. Recently, \cite{sunberg2017continuous} presented an online algorithm, POMCPOW, for continuous state, actions and observations which can be applied to BAMDP problems. Of course, online and offline approaches can be combined, e.g. by using the offline policy as a default rollout policy.

The aforementioned approaches aim for asymptotic guarantees. On the other hand, PAC-MDP~\cite{strehl2009reinforcement} approaches seek to bound the number of exploration steps before achieving near-optimal performance. This was originally formulated in the context of discrete MDPs with unknown transition and reward functions~\cite{brafman2002r,strehl2006pac} and extended to continuous spaces~\cite{kakade2003exploration,pazis2013pac}. BOSS~\cite{asmuth2009bayesian} first introduced the notion of uncertainty over model parameters, albeit for a PAC-MDP style guarantee. The \pacbayes property for an MDP was formally introduced in ~\cite{kolter2009near}, as discussed in the previous subsection.

There are several effective heuristic-based approaches~\cite{dearden1998bayesian,strens2000bayesian} to BAMDP that we omit for brevity. We refer the reader to~\cite{ghavamzadeh2015bayesian} for a comprehensive survey. We also compare with QMDP~\cite{littman1995learning} which approximates the expected Q-value with respect to the current belief and greedily chooses an action.

\begin{table}[!t]
\small
\centering
\begin{tabulary}{1.0\columnwidth}{RCCCC}\toprule
  \multirow{2}{*}{\bf Algorithm}                    & {\bf Continuous}   & \multirow{2}{*}{\bf PAC} & \multirow{2}{*}{\bf Offline} \\
                                                    & {\bf State/Action} &                          &                              \\ \midrule
  SARSOP~\shortcite{kurniawati2008sarsop}           & \xmark             & \xmark                   & \cmark \\
  \algPOMDPLite~\shortcite{chen2016pomdp}           & \xmark             & \cmark                   & \xmark \\
  POMCPOW~\shortcite{sunberg2017continuous}         & \cmark             & \xmark                   & \xmark \\   \arrayrulecolor{white}\midrule
  \algName~(Us)                                     & \cmark             & \cmark                   & \cmark \\
  \arrayrulecolor{black} \bottomrule
\end{tabulary}
\caption{Comparison of BAMDP algorithms}
\label{tab:benchmark}
\end{table}

\tabref{tab:benchmark} compares the key features of \algName against a selection of prior work.


\section{\algName: Continuous PAC Optimal Exploration in Belief Space}\label{sec:approach}

In this section, we present \algName, an offline \pacbelief algorithm that computes a near-optimal policy for a continuous state and action BAMDP.
\algName is an extension of \algCPACE~\cite{pazis2013pac}, a PAC optimal algorithm for continuous state and action MDPs.
Efficient exploration of a continuous space is challenging because that the same state-action pair cannot be visited more than once.
\algCPACE addresses this by assuming that the state-action value function is Lipschitz continuous, allowing the value of a state-action pair to be approximated with nearby samples.
Similar to other PAC optimal algorithms~\cite{strehl2009reinforcement}, \algCPACE applies the principle of \emph{optimism in the face of uncertainty}:
the value of a state-action pair is approximated by averaging the value of nearby samples, inflated proportionally to their distances.
Intuitively, this distance-dependent bonus term encourages exploration of regions that are far from previous samples until the optimistic estimate results in a near-optimal policy.

Our key insight is that \algCPACE can be extended from continuous states to those augmented with finite-dimensional belief states.
We derive sufficient conditions for Lipschitz continuity of the belief value function.
We show that \algName is indeed \pacbelief and bound the sample complexity as a function of the covering number of the reachable belief space from initial belief $\belief_0$.
In addition, we also present and analyze three practical strategies for improving the sample complexity and runtime of \algName.

\subsection{Definitions and Assumptions}

We assume all rewards lie in $[0, \Rmax]$ which implies $0 \leq \Qmax, \Vmax \leq \frac{\Rmax}{1 - \gamma}$.
We will first show that \assref{assump:smooth_reward_transition} and \assref{assump:belief_contraction} are sufficient conditions for Lipschitz continuity of the value function.\footnote{For all proofs, refer to \supp.}
Subsequent proofs do not depend on these assumptions as long as the value function is Lipschitz continuous.

\begin{assumption}[Lipschitz Continuous Reward and Transition Functions]
\label{assump:smooth_reward_transition}
Given any two state-action pairs $\left(\state_1, \action_1\right)$ and $\left(\state_2, \action_2\right)$, there exists a distance metric $\dis{\cdot}{\cdot}$ and Lipschitz constants $\LR, \LP$ such that the following is true:
\begin{equation*}
\begin{aligned}
\abs{ \rewardFn(\state_1,\latent, \action_1) -  \rewardFn(\state_2, \latent, \action_2)} \leq \LR d_{\state_1,\action_1,\state_2,\action_2} \\
\sum_{\state'}  \abs{P(\state' | \state_1, \latent, \action_1) - P(\state' | \state_2, \latent, \action_2)} \leq \LP d_{\state_1,\action_1,\state_2,\action_2} \\
\end{aligned}
\end{equation*}
where $d_{\state_1,\action_1,\state_2,\action_2} = \dis{(\state_1,\action_1)}{(\state_2,\action_2)}$

\end{assumption}

\begin{assumption}[Belief Contraction]
\label{assump:belief_contraction}
Given any two belief vectors $\belief_1, \belief_2$ and any tuple of $(\state, \action, \state')$, the updated beliefs from the Bayes estimator $\belief'_1 = \estimator(\belief_1, \state, \action, \state')$ and $\belief'_2 = \estimator(\belief_2, \state, \action, \state')$ satisfy the following:
\begin{equation*}
\norm{\belief'_1 - \belief'_2}{1} \leq \norm{\belief_1 - \belief_2}{1}
\end{equation*}
\end{assumption}

\assref{assump:smooth_reward_transition} and \assref{assump:belief_contraction} can be used to prove the following lemma.
\begin{lemma}[Lipschitz Continuous Value Function]
\label{lemma:smooth_action_value}
Given any two state-belief-action tuples $(\state_1,\belief_1,\action_1)$ and $(\state_2,\belief_2,\action_2)$, there exists a distance metric $\dis{\cdot}{\cdot}$ and a Lipschitz constant $\LQ$ such that the following is true:
\begin{equation*}
\begin{aligned}
| \Q(\state_1, \belief_1, \action_1) - \Q(\state_2, \belief_2, & \action_2) |   \leq \LQ d_{\state_1,\belief_1,\action_1,\state_2,\belief_2,\action_2}
\end{aligned}
\end{equation*}
where $d_{\state_1,\belief_1,\action_1,\state_2,\belief_2,\action_2} = \dis{(\state_1,\belief_1, \action_1)}{(\state_2,\belief_2, \action_2)}$
\end{lemma}
The distance metric $\dis{(\state_1,\belief_1, \action_1)}{(\state_2,\belief_2, \action_2)}$ for state-belief-action tuples is a linear combination of the distance metric for state-action pairs used in \assref{assump:smooth_reward_transition} and the $L_1$ norm for belief
\begin{equation*}
\begin{aligned}
\alpha \dis{(\state_1, \action_1)}{(\state_2, \action_2)} &+ \norm{\belief_1 - \belief_2}{1}
\end{aligned}
\end{equation*}
for an appropriate choice of $\alpha$, which is a function of $R_{\max}, L_R,$ and $L_P$.

\algName builds an optimistic estimator $\tQ(\state, \belief, \action)$ for the value function $\Q(\state, \belief, \action)$ using nearest neighbor function approximation from a collected sample set.
Since the value function is Lipschitz continuous, the value for any query can be estimated by extrapolating the value of neighboring samples with a distance-dependent bonus.
If the number of close neighbors is sufficiently large, the query is said to be ``known'' and the estimate can be bounded.
Otherwise, the query is unknown and is added to the sample set.
Once enough samples are added, the entire reachable space will be known and the estimate will be bounded with respect to the true optimal value function $\Q^*(\state, \belief, \action)$.
We define these terms more formally below.

\begin{definition}[Known Query]
\label{def:known_query}
Let $\LtQ = 2 \LQ$ be the Lipschitz constant of the optimistic estimator. A state-belief-action query $(\state, \belief, \action)$ is said to be "known" if its $k\ts{th}$ nearest neighbor in the sample set $(\state_k, \belief_k, \action_k)$ is within $\epsD / \LtQ$.
\end{definition}

We are now ready to define the estimator.
\begin{definition}[Optimistic Value Estimate]
\label{def:optimistic_estimate}
Assume we have a set of samples $\sampleSet$ where every element is a tuple $(\state_i, \belief_i, \action_i, \reward_i, \state'_i, \belief'_i)$: starting from $(\state_i, \belief_i)$, the agent took an action $\action_i$, received a reward $\reward_i$, and transitioned to $(\state'_i, \belief'_i)$. Given a state-belief-action query $(\state,\belief,\action)$, its $j\ts{th}$ nearest neighbor from the sample set provides an optimistic estimate
\begin{equation}
  \estimate_j = \LtQ \dis{(\state,\belief,\action)}{(\state_j,\belief_j,\action_j)} + \tQ(\state_j, \belief_j, \action_j).
\end{equation}
The value is the average of all the nearest neighbor estimates
\begin{equation}
\label{eq:optimisic_bellman_operator}
\tQ(\state, \belief, \action) = \frac{1}{k} \sum_{j=1}^k  \min \left( \estimate_j, \tQmax \right)
\end{equation}
where $\tQmax = \Rmax + \gamma \Qmax$ is the upper bound of the estimate. If there are fewer than $k$ neighbors, $\tQmax$ can be used in place of the corresponding $\estimate_j$.
\end{definition}

Note that the estimator is a recursive function. Given a sample set $\sampleSet$, value iteration is performed to compute the estimate for each of the sample points,
\begin{equation}
 \tQ(\state_i, \belief_i, \action_i) = r_i + \gamma \max_\action \tQ(\state_i', \belief_i', \action)
\end{equation}
where $\tQ(\state_i', \belief_i', \action)$ is approximated via \eqref{eq:optimisic_bellman_operator} using its nearby samples. This estimate must be updated every time a new sample is added to the set.

We introduce two additional techniques that leverage the Q-values of the underlying latent MDPs to improve the sample complexity of \algName.

\begin{definition}[Best-Case Upper Bound]
\label{def:heuristic_upper_bound}
We can replace the constant $\tQmax$ in \defref{def:optimistic_estimate} with $\tQmax(\state, \belief, \action)$ computed as follows:
\begin{equation*}
    \tQmax(\state, \belief, \action) = \max_{\latent, \belief(\latent) > 0} \Q(\state, \latent, \action)
\end{equation*}
\end{definition}
In general, any admissible heuristic $U$ that satisfies $Q(\state, \belief, \action) \leq U(\state,\belief,\action) \leq \tQ_{\max}$ can be used.
In practice, the Best-Case Upper Bound reduces exploration of actions which are suboptimal in all latent MDPs with nonzero probability.

We can also take advantage of $Q(\state, \latent, \action)$ whenever the belief distribution collapses.
These exact values for the latent MDPs can be used to seed the initial estimates.
\begin{definition}[Known Latent Initialization]
\label{def:seeding_exact}
Let $\onehot{\phi}$ be the belief distribution where $P(\phi) = 1$, i.e. a one-hot encoding.
If there exists $\phi$ such that $\norm{\belief - \onehot{\phi}}{1} \leq \frac{\epsilon}{\LQ(1 + \gamma)}$, then we can use the following estimate:
\begin{equation}
\label{eq:estimate_exact_seed}
  \tQ(\state, \belief, \action) = \Q(\state,\latent,\action) 
\end{equation}
This extends \defref{def:known_query} for a known query to include any state-belief-action tuple where the belief is within $\frac{\epsilon}{\LQ(1 + \gamma)}$ of a one-hot vector.
\end{definition}
We refer to \propref{prop:seeding} for how this reduces sample complexity.

\subsection{Algorithm}

\begin{algorithm}[!t]
\caption{\algName}
\label{alg:beliefCPACE}
\begin{algorithmic}[1]
\Require Bayes-Estimator $\estimator$, initial belief $\belief_0$, BAMDP $\BAMDP$,\par terminal condition $G$, horizon $\hor$
\Ensure Action value estimate $\tQ$
\Statex
\State Initialize sample set $\sampleSet \gets \emptyset$
\While {$G$ is false}
  \State Initialize $\BAMDP$ by resampling initial state and latent \par variable to $\state_0, \latent_0 \sim P_0(\state, \latent)$
  \State Reset belief to $\belief_0$
  \For{$t = 0, 1, 2 , \cdots, \hor-1$} \label{alg:collect}
    \State Compute action $\action_t \gets \arg\max_{\action} \tQ (\state_t, \belief_t, \action)$ \label{alg:beliefCPACE:greedy}
    \State Execute $\action_t$ on $\BAMDP$ to receive $\reward_t, \state_{t+1}$
    \State Invoke $\estimator$ to get $\belief_{t+1} \gets \estimator(\belief_t, \state_t, \action_t, \state_{t+1})$ \label{alg:beliefCPACE:belief}
    \If{ $\left(\state_t, \belief_t, \action_t\right)$ is not known}
      \State Add $(\state_t, \action_t, \belief_t, \reward_t, \state_{t+1}, \belief_{t+1})$ to $\sampleSet$ \label{alg:beliefCPACE:add}
      \State Find fixed point of $\tilde{Q}(\state_i, \belief_i, \action_i)$ for $\sampleSet$ \label{alg:beliefCPACE:fixed}
    \EndIf
  \EndFor \label{alg:collect-end}
\EndWhile
\State \textbf{Return} $\tQ$
\Statex
\Function{$\tQ(\state, \belief, \action)$}{} \label{alg:beliefCPACE:funcStart}
\State Find closest one-hot vector $\phi = \min_{\phi} \dis{\belief}{\onehot{\phi}}$
\If{$\norm{\belief - \onehot{\phi}}{1} \leq \frac{\epsilon}{\LQ(1 + \gamma)}$}
  \State $\tQ(\state, \belief, \action) \gets \Q(\state,\phi,\action)$
\Else
\State Find $k$ nearest neighbors $\{\state_j, \belief_j, \action_j, \reward_j, \state'_j, \belief'_j \}$ \par \hskip\algorithmicindent in sample set $\sampleSet$
\For{$j = 1, \cdots, k$}
  \State $d_j \gets \dis{(\state,\belief,\action)}{(\state_j,\belief_j,\action_j)}$
  \State $\estimate_j \gets \LtQ d_j + \tQ(\state_j, \belief_j, \action_j)$
\EndFor
\State $\tQ(\state, \belief, \action) = \frac{1}{k} \sum_{j=1}^k  \min \left( \estimate_j, \tQmax(\state, \belief, \action) \right)$
\EndIf
\State \textbf{Return} $\tQ(\state, \belief, \action)$ \label{alg:beliefCPACE:funcEnd}
\EndFunction
\end{algorithmic}
\end{algorithm}

We describe our algorithm, \algName, in \aref{alg:beliefCPACE}. To summarize, at every timestep $t$ the algorithm computes a greedy action $\action_t$ using its current value estimate $\tQ(\state_t, \belief_t, \action_t)$, receives a reward $\reward_t$, and transitions to a new state-belief $(\state_{t+1}, \belief_{t+1})$~\linesref{alg:beliefCPACE:greedy}{alg:beliefCPACE:belief}. If the sample is not known, it is added to the sample set $\sampleSet$ \lineref{alg:beliefCPACE:add}. The value estimates for all samples are updated until the fixed point is reached~\lineref{alg:beliefCPACE:fixed}. Terminal condition $G$ is met when no more samples are added and value iteration has converged for sufficient number of iterations. The algorithm invokes a subroutine for computing the estimated value function \linesref{alg:beliefCPACE:funcStart}{alg:beliefCPACE:funcEnd} which correspond to the operations described in \defref{def:optimistic_estimate},~\ref{def:heuristic_upper_bound}, and ~\ref{def:seeding_exact}.

\subsection{Analysis of Sample Complexity}

We now prove that \algName is \pacbelief. Since we adopt the proof of \algCPACE, we only state the main steps and defer the full proof to \supp.
We begin with the concept of a known belief MDP.
\begin{definition}[Known Belief MDP]
\label{def:known-bmdp}
Let $\beliefMDP$ be the original belief MDP. Let $\knownSet$ be the set of all known state-belief-action tuples. We define a known belief MDP $\knownBeliefMDP$ that is identical to $\beliefMDP$ on $\knownSet$ (i.e. identical transition and reward functions) and for all other state-belief-action tuples, it transitions deterministically with a reward $\rewardFn(\state, \belief, \action) = \tQ(\state, \belief, \action)$ to an absorbing state with zero reward.
\end{definition}

We can then bound the performance of a policy on $\beliefMDP$ with its performance on $\knownBeliefMDP$ and the maximum penalty incurred by escaping it.

\begin{lemma}[Generalized Induced Inequality, Lemma 8 in \cite{strehl2008online}]
\label{lemma:generalized-induced-inequality}
We are given the original belief MDP $\beliefMDP$, the known belief MDP $\knownBeliefMDP$, a policy $\policy$ and time horizon $\hor$. Let $P(\escapeEvent)$ be the probability of an escape event, i.e. the probability of sampling a state-belief-action tuple that is not in $\knownSet$ when executing $\policy$ on $\beliefMDP$ from $(\state, \belief)$ for $\hor$ steps. Let $\V^{\policy}_{\beliefMDP}$ be the value of executing policy on $\beliefMDP$. Then the following is true:
\begin{equation*}
  \V^{\policy}_{\beliefMDP}(\state, \belief, \hor) \geq \V^{\policy}_{\knownBeliefMDP}(\state, \belief, \hor) - \Qmax P (\escapeEvent)
\end{equation*}
\end{lemma}

We now show one of two things can happen: either the greedy policy escapes from the known MDP, or it remains in it and performs near optimally. We first show that it can only escape a certain number of times before the entire reachable space is known.

\begin{lemma}[Full Coverage of Known Space, Lemma 4.5 in \cite{kakade2003exploration}]
\label{lemma:kakade-known}
All reachable state-belief-action queries will become known after adding at most $k \cover(\epsD/\LtQ)$ samples to $\sampleSet$.
\end{lemma}

\begin{corollary}[Bounded Escape Probability]
\label{corollary:escape}
At a given timestep, let $P(\escapeEvent) > \frac{\epsilon}{\Qmax(1-\gamma)}$. Then with probability $1 - \delta$, this can happen at most for $\frac{2 \Qmax}{\epsilon} \left( k \cover(\epsD/\LtQ) + \log\left(\frac{1}{\delta}\right) \right)\log \frac{\Rmax}{\epsilon}$ timesteps.
\end{corollary}

We now show that when inside the known MDP, the greedy policy will be near optimal.

\begin{lemma}[Near-optimality of Approximate Greedy (Theorem 3.12 of \cite{pazis2013pac})]
\label{lemma:near-optimality}
Let $\tQ$ be an estimate of the value function that has bounded Bellman error $-\epsilon_{-} \leq \tQ - B \tQ \leq \epsilon_{+}$, where $B$ is the Bellman operator. Let $\tpolicy$ be the greedy policy on $\tQ$. Then the policy is near-optimal:
\begin{equation*}
  \V^{\tpolicy}(\state, \belief) \geq \V^*(\state, \belief) - \frac{\epsilon_{-} + \epsilon_{+}}{1 - \gamma}
\end{equation*}
\end{lemma}

Let $\epsilon$ be the approximation error caused by using a finite number of neighbors in \eqref{eq:optimisic_bellman_operator} instead of the Bellman operator. Then \lemref{lemma:near-optimality} leads to the following corollary.

\begin{corollary}[Near-optimality on Known Belief MDP]
\label{corollary:near-optimality-known-BMDP}
If $\frac{\tQmax^2}{\epsilon^2} \log \left(\frac{2 \cover(\epsD/\LtQ) }{ \delta } \right) \leq k \leq \frac{2 \cover(\epsD/\LtQ)}{\delta}$, i.e. the number of neighbors is large enough, then using Hoeffding's inequality we can show $-\epsilon \leq \tQ - B\tQ \leq 2\epsilon$. Then on the known belief MDP $\knownBeliefMDP$, the following can be shown with probability $1-\delta$:
\begin{equation*}
  \V^{\tpolicy}_{\knownBeliefMDP}(\state, \belief) \geq \V_{\knownBeliefMDP}^*(\state, \belief) - \frac{3\epsilon}{1 - \gamma}
\end{equation*}
\end{corollary}

We now put together these ideas to state the main theorem.

\begin{theorem}[\algName is \pacbelief]
\label{theorem:belief_cpace_pac}
Let $\beliefMDP$ be a belief MDP. At timestep $t$, let $\tpolicy_t$ be the greedy policy on $\tQ$, and let $(\state_t, \belief_t)$ be the state-belief pair. With probability at least $1-\delta$, $\V^{\tpolicy_t}(\state_t, \belief_t) \geq \V^*(\state_t, \belief_t) - \frac{7\epsilon}{1 - \gamma}$, i.e. the algorithm is $\frac{7\epsilon}{1 - \gamma}$-close to the optimal policy for all but
\begin{equation*}
\sample = \frac{2 \Qmax}{\epsilon}  \left( k \cover(\epsilon/\LtQ) + \log \frac{2}{\delta}  \right)\log \left( \frac{\Rmax}{\epsilon} \right)
\end{equation*}
steps when $k \in \left[\frac{\tQmax^2}{\epsilon^2} \log  \frac{4 \cover(\epsilon/\LtQ) }{ \delta } , \frac{4 \cover(\epsilon/\LtQ)}{\delta}\right]$ is used for the number of neighbors in \eqref{eq:optimisic_bellman_operator}.
\end{theorem}
\begin{proofsketch} At time $t$, we can form a known belief MDP $\knownBeliefMDP$ from the samples collected so far. Either the policy leads to an escape event within the next $T$ steps or the agent stays within $\knownBeliefMDP$. Such an escape can happen at most $m$ times with high probability; when the escape probability is low, $V^{\tpolicy}$ is $\frac{7\epsilon}{1 - \gamma}$-optimal.
\end{proofsketch}

\subsection{Analysis of Performance Enhancements}
We can initialize estimates with exact Q values for the latent MDPs.
This makes the known space larger, thus reducing covering number.
\begin{proposition}[Known Latent Initialization]
\label{prop:seeding}
Let $\coverReduced(\epsD/\LtQ)$ be the covering number of the reduced space $\setst{(\state, \belief, \action)}{\forall \onehot{i},  \dis{\belief}{\onehot{i}} \geq \frac{\epsilon}{\LQ(1 + \gamma)}}$. Then the sample complexity reduces by a factor of $\frac{\coverReduced(\epsD/\LtQ)}{\cover(\epsD/\LtQ)}$.
\end{proposition}

It is also unnecessary to perform value iteration until convergence.
\begin{proposition}[Approximate Value Iteration]
\label{prop:approx_vi}
Let $0 < \tolVI < \tQmax$. Suppose the value iteration step \lineref{alg:beliefCPACE:fixed} is run only for $i = \lceil \nicefrac{ \log \left( \nicefrac{\tolVI}{\tQmax} \right) }{\log \gamma} \rceil$ iterations denoted by $\tB^i \tQ$ (instead of until convergence $\tB^\infty \tQ$). We can bound the difference between two functions as $\norm{ \tB^i \tQ - \tB^\infty \tQ }{\infty} \leq \tolVI$. This results in an added suboptimality term in \thmref{theorem:belief_cpace_pac}:
\begin{equation}
\V^{\tpolicy_t}(\state_t, \belief_t) \geq \V^*(\state_t, \belief_t) - \frac{7\epsilon + 2 \tolVI}{1 - \gamma}
\end{equation}
\end{proposition}

One practical enhancement is to collect new samples in a batch with a fixed policy before performing value iteration.
This requires two changes to the algorithm: 1) an additional loop to repeat \linesref{alg:collect}{alg:collect-end} $n$ times, and 2) perform \lineref{alg:beliefCPACE:fixed} outside of the loop.
This increases the sample complexity by a constant factor but has empirically reduced runtime by only performing value iteration when a large change is expected.
\begin{proposition}[Batch Sample Update]
\label{prop:batch-update}
Suppose we collect new samples from $n$ rollouts with the greedy policy at time $t$ before performing value iteration.
This increases the sample complexity only by a constant factor of $O(n)$.
\end{proposition}


\begin{figure*}[!t]
  \centering
  \begin{subfigure}[b]{0.47\linewidth}
  \includegraphics[width=0.9\textwidth]{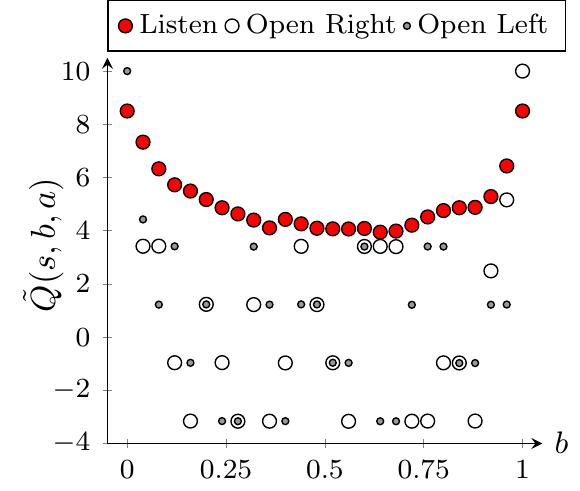}
  \caption{
  Value approximation for Tiger
  }
  \label{fig:tiger-belief}
  \end{subfigure}
  \hfill
  \begin{minipage}[b]{0.49\linewidth}
  \begin{subfigure}[b]{1\linewidth}
  \centering
    \includegraphics[width=0.8\textwidth]{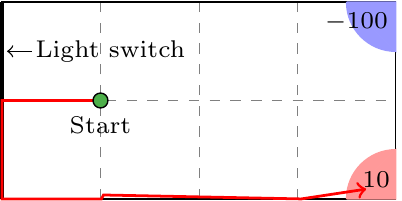}
    \caption{One optimal path taken by $\algName$ for Light-Dark Tiger.
    }
    \label{fig:lightdarktiger}
  \end{subfigure}

  \vspace{0.5em}

  \begin{subfigure}[b]{1\linewidth}
  \footnotesize
  \begin{tabulary}{1\textwidth}{LCCCC}\toprule
                          & {\bf \algQMDP}        & {\bf \algPOMDPLiteShort}   & {\bf \algSARSOP}     & {\bf \algNameShort} \\ \midrule
  \envTiger               & $16.5 \pm .8$      & $11.8 \pm .6$       &$\bf17.8 \pm 1.9$   & $\bf 18.0 \pm 1.4$\\
  \envChain               & $12.9 \pm .5$      &  $13.0 \pm .1$      &$13.4 \pm  .1 $     & $\bf 14.3  \pm .1 $\\
  \envLightDark           & $0$                & $15.1 \pm .3$        &    $ \bf 29.0 $    & $\bf 29.0$\\
  \rowcolor{lightgreen}\envLightDark (cont.)   & $0$                  & -                   &       -            & $\bf 25.4 \pm .1$\\
  \bottomrule
  \end{tabulary}
  \caption{Benchmark results. \envLightDark (cont.) has continuous state space.
    }
  \label{tab:benchmark_results}
  \end{subfigure}
   \end{minipage}
   \caption{With greedy exploration, only best actions are tightly approximated~(\figref{fig:tiger-belief}). \algName takes optimal actions for a continuous BAMDP~(\figref{fig:lightdarktiger}). \algName is competitive for both discrete and continuous BAMDPs~(\tabref{tab:benchmark_results}).}
   \vspace{-1em}
\end{figure*}

\section{Experimental Results}
\label{sec:results}
We compare \algName with \algQMDP, \algPOMDPLite, and \algSARSOP for discrete BAMDPs and with \algQMDP for continuous BAMDPs.
For discrete state spaces, we evaluate \algName on two widely used synthetic examples, Tiger~\cite{kaelbling1998planning} and Chain~\cite{strens2000bayesian}. For both \algName and \algPOMDPLite, the parameters were tuned offline for best performance.
For continuous state spaces, we evaluate on a variant of the Light-Dark problem~\cite{platt2010belief}.

While our analysis is applicable for BAMDPs with continuous state and action spaces, any approximation the greedy selection of an action is not guaranteed to be \pacbayes. Thus, we limit our continuous BAMDP experiments to discrete action spaces and leave the continuous action case for future work.

\textbf{Tiger:}
We start with the Tiger problem. The agent stands in front of two closed doors and can choose one of three actions: listen, open the left door, or open the right door.
One of the doors conceals a tiger; opening this door results in a penalty of -100, while the other results in a reward of 10.
Listening informs the agent of the correct location of the tiger with probability $0.85$, with a cost of -1.
As observed by \cite{chen2016pomdp}, this POMDP problem can be cast as a BAMDP problem with two latent MDPs.

\tabref{tab:benchmark_results} shows that \algName performs as competitively as \algSARSOP and is better than \algQMDP or \algPOMDPLite.
This is not surprising since both \algName and \algSARSOP are offline solvers.

\figref{fig:tiger-belief} visualizes the estimated values.
Because \algName explores greedily, exploration is focused on actions with high estimated value, either due to optimism from under-exploration or actual high value.
As a result, suboptimal actions are not taken once \algName is confident that they have lower value than other actions.
Because fewer samples have been observed for these suboptimal actions, their approximated values are not tight. Note also that the original problem explores a much smaller subset of the belief space, so we have randomly initialized the initial belief from $[0, 1]$ rather than always initializing to 0.5 for this visualization, forcing $\algName$ to perform additional exploration.

\textbf{Chain: }
The Chain problem consists of five states $\{s_i\}_{i=1}^5$ and two actions $\{A, B\}$.
Taking action $A$ in state $s_i$ transitions to $s_{i+1}$ with no reward; taking action $A$ in state $s_5$ transitions to $s_5$ with a reward of 10.
Action $B$ transitions from any state to $s_1$ with a reward of 2.
However, these actions are noisy: in the canonical version of Chain, the opposite action is taken with slip probability 0.2.
In our variant, we allow the slip probability to be selected from $[0.2, 0.5, 0.8]$ with uniform probability at the beginning of each episode.
These three latent MDPs form a BAMDP.
\tabref{tab:benchmark_results} shows that \algName outperforms other algorithms.

\textbf{Light-Dark Tiger: }
We consider a variant of the Light-Dark problem, which we call Light-Dark Tiger~(\figref{fig:lightdarktiger}). In this problem, one of the two goal corners (top-right or bottom-right) contains a tiger. The agent receives a penalty of -100 if it enters the goal corner containing the tiger and a reward of 10 if it enters the other region.
There are four actions---Up, Down, Left, Right---which move one unit with Gaussian noise of $\mathcal{N}(0, \sigma^2)$.
The tiger location is unknown to the agent until the left wall is reached. As in the original Tiger problem, this POMDP can be formulated as a BAMDP with two latent MDPs.

We consider two cases, one with zero noise and another with $\sigma = 0.01$. With zero noise, the problem is a discrete POMDP and the optimal solution is deterministic; the agent hits the left wall and goes straight to the goal location. When there is noise, the agent may not reach the left wall in the first step. Paths executed by \algName still take Left until the left wall is hit and goes to the goal~(\figref{fig:lightdarktiger}).


\section{Discussion}
\label{sec:conclusion}

We have presented the first \pacbayes algorithm for continuous BAMDPs whose value functions are Lipschitz continuous. While the practical implementation of \algName is limited to discrete actions, our analysis holds for both continuous and discrete state and actions. We believe that our analysis provides an important insight for the development of PAC efficient algorithms for continuous BAMDPs.

The BAMDP formulation is useful for real-world robotics problems where uncertainty over latent models is expected at test time. An efficient policy search algorithm must incorporate prior knowledge over the latent MDPs to take advantage of this formulation. As a step toward this direction, we have introduced several techniques that utilize the value functions of underlying latent MDPs without affecting PAC optimality.

One of the key assumptions \algName has made is that the cardinality of the latent state space is finite. This may not be true in many robotics applications in which latent variables are drawn from continuous distributions. In such cases, the true BAMDP can be approximated by sampling a set of latent variables, as introduced in ~\cite{wang2012monte}. In future work, we will investigate methods to select representative MDPs and to bound the gap between the optimal value function of the true BAMDP and the approximated one.

Although it is beyond the scope of this paper, we would like to make two remarks. First, \algName can easily be extended to allow parallel exploration, similar to how \cite{pazis2016efficient} extended the original C-PACE to concurrently explore multiple MDPs. Second, since we have generative models for the latent MDPs, we may enforce exploration from arbitrary belief points. Of course, the key to efficient exploration of belief space lies in exploring just beyond the optimally reachable belief space, so ``random'' initialization is unlikely to be helpful. However, if we can approximate this space similarly to sampling-based kinodynamic planning algorithms \cite{yanbo2016}, this may lead to more structured search in belief space.



\section{Acknowledgements}

 This work was partially funded by Kwanjeong Educational Foundation, NASA Space Technology Research Fellowships (NSTRF), the National Institute of Health R01 (\#R01EB019335), National Science Foundation CPS (\#1544797), National Science Foundation NRI (\#1637748), the Office of Naval Research, the RCTA, Amazon, and Honda.


\bibliography{main}  
{
\tiny
\bibliographystyle{aaai}
}
\clearpage

\clearpage

\section{Supplementary Material}


\subsection{Proof of Lemma~\ref{lemma:smooth_action_value}}

The proof has a few key components. Firstly, we show that the reward and transition functions are Lipschitz continuous. Secondly, we show that the Q value that differ only in belief is Lipschitz continuous. Finally, we put these together to show that the Q value in state-belief-action space is Lipschitz continuous.
For notational simplicity, let $\stateaction := (\state, \action) $.

\bigskip
\noindent\textbf{Lipschitz continuity for reward and transition functions}

\vspace{1em}
\noindent We begin by showing that the reward as a function of the state-belief-action is Lipschitz continuous. For any two tuples $(\stateaction_1, \belief_1)$ and $(\stateaction_2, \belief_2)$, the following is true:
\begin{equation}
\begin{aligned}
	\label{eq:rew_lipschitz_bound}
    &\abs{ \rewardFn(\stateaction_1, \belief_1) -  \rewardFn(\stateaction_2, \belief_2)}  \\
    &\leq \sum\nolimits_\latent \abs{ \rewardFn(\stateaction_1, \latent) \belief_1(\latent) - \rewardFn(\stateaction_2, \latent) \belief_2(\latent) }  \\
    &\leq \sum\nolimits_\latent | \rewardFn ( \stateaction_1, \latent) \belief_1(\latent) - \rewardFn( \stateaction_2, \latent) \belief_1(\latent)  \\
    &\quad  + \rewardFn(\stateaction_2, \latent)\belief_1(\latent) -  \rewardFn(\stateaction_2, \latent) \belief_2(\latent) |  \\
    & \leq \sum\nolimits_\latent \abs{ \rewardFn(\stateaction_1, \latent) - \rewardFn(\stateaction_2,\latent) } \belief_1(\latent)  \\
    &\quad +  \sum\nolimits_\latent \rewardFn(\stateaction_2, \latent) \abs{ \belief_1(\latent) - \belief_2(\latent)}   \\
    &\leq \sum\nolimits_\latent \LR d_{\stateaction_1,\stateaction_2} b_1(\latent) + \Rmax \norm{ \belief_1 - \belief_2 }{1} \\
    &\leq \LR d_{\stateaction_1,\stateaction_2} + R_{\max} \norm{ \belief_1 - \belief_2 }{1}
\end{aligned}
\end{equation}
where we have used Assumption~\ref{assump:smooth_reward_transition} for the 4th inequality.

Similarly, the state transition as a function of the state-belief-action can also be shown to be Lipschitz continuous:

\begin{equation}
\begin{aligned}
	\label{eq:prob_lipschitz_bound}
    &\sum\nolimits_{\state'} \abs{ \left( P(\state'|\stateaction_1,\belief_1) - P(\state'|\stateaction_2,\belief_2) \right)} \\
    &\leq \sum\nolimits_{\state',\latent} \abs{  P(\state' | \stateaction_1, \latent) \belief_1(\latent) - P(\state' | \stateaction_2, \latent) \belief_2(\latent) }\\
    &\leq \sum\nolimits_{\state',\latent}  | P ( \state' | \stateaction_1, \latent) \belief_1(\latent) - P( \state' | \stateaction_2, \latent) \belief_1(\latent) \\
    &\quad + P( \state' | \stateaction_2, \latent) \belief_1(\latent) -  P( \state' | \stateaction_2, \latent) \belief_2(\latent) | \\
    &\leq \sum\nolimits_{\state',\latent} \abs{ P( \state' | \stateaction_1,\latent) - P(\state' | \stateaction_2,\latent) } \belief_1(\latent) \\
    &\quad  + P( \state' | \stateaction_2, \latent) \abs{ \belief_1(\latent) - \belief_2(\latent)}  \\
    &\leq \LP  d_{\stateaction_1,\stateaction_2} + \norm{ \belief_1 - \belief_2 }{1}
\end{aligned}
\end{equation}
where we have used Assumption~\ref{assump:smooth_reward_transition} for the 4th inequality.


\vspace{1em}
\noindent\textbf{Lipschitz continuity for fixed state-action Q value}
\vspace{1em}

We'll use the following inequality. For two positive bounded functions $f(x)$ and $g(x)$,
\begin{equation}\label{eq:max_func_diff}
\abs{ \max_x f(x) - \max_x g(x) } \leq \max_x \abs{f(x) - g(x)}
\end{equation}

\noindent First let's assume the following is true:
\begin{equation} \label{eq:q_belief_temp}
    \| \Q(\stateaction, \belief_1) -  \Q(\stateaction, \belief_2)\|  \leq \LQB \norm{ \belief_1 - \belief_2 }{1}
\end{equation}
We will derive the value of $\LQB$ (if it exists) by expanding the expression for the action value function.

Let $\belief' = \estimator(\belief, \state, \action, \state')$ be the deterministic belief update. We have the following:
\begin{equation}
\begin{aligned} \label{eq:q-continuous}
    & \abs{ \Q(\stateaction, \belief_1) - \Q(\stateaction, \belief_2)}  \\
    & \leq | \rewardFn(\stateaction, \belief_1) - \rewardFn(\stateaction, \belief_2)  \\
    & \quad + \gamma \sum\nolimits_{\state'} P(\state'| \stateaction, \belief_1) \V(\state', \belief'_1) - P(\state' | \stateaction, \belief_2) \V(\state', \belief'_2) |  \\
    & \leq \Rmax \norm{ \belief_1 - \belief_2 }{1} \\
    & \quad + \gamma  \sum\nolimits_{\state'} \abs{ \left( P(\state'|\stateaction,\belief_1) - P(\state'|\stateaction,\belief_2) \right) \V(\state',\belief_1') }   \\
    & \quad + \gamma  \sum\nolimits_{\state'} P(\state'|\stateaction,\belief_2) \abs{ \V(\state',\belief_1') - \V(\state',\belief'_2)  } \\
    & \leq \Rmax \norm{ \belief_1 - \belief_2 }{1}  + \gamma \Vmax \norm{ \belief_1 - \belief_2 }{1}   \\
    & \quad + \gamma  \sum\nolimits_{\state'} P(\state'|\stateaction,\belief_2) \abs{ \max_{\action'} \Q(\stateaction', \belief'_1) - \max_{\action'} \Q(\stateaction',\belief'_2)   }  \\
    & \leq (\Rmax + \gamma \Vmax) \norm{ \belief_1 - \belief_2 }{1}\\
    &\quad+ \gamma  \sum\nolimits_{\state'} P(\state'|\stateaction,\belief_2) \max_{\action'} \abs{ \Q(\stateaction', \belief'_1) - \Q(\stateaction',\belief'_2) }  \\
    & \leq (\Rmax + \gamma \Vmax) \norm{ \belief_1 - \belief_2 }{1} + \gamma  \LQB \norm{ \belief'_1 - \belief'_2 }{1}\\ 
    & \leq \left( \Rmax + \gamma (\Vmax + \LQB) \right) \norm{ \belief_1 - \belief_2 }{1}
\end{aligned}
\end{equation}
where we have used \eref{eq:rew_lipschitz_bound}, \eref{eq:prob_lipschitz_bound}, \eref{eq:max_func_diff}, \eref{eq:q_belief_temp}, and Assumption~\ref{assump:belief_contraction} for the 2nd, 3rd, 4th, 5th and last inequalities, respectively.

\vspace{1em}
Applying above inequality to \eref{eq:q_belief_temp}, we can solve for $\LQB$:
\begin{equation}
\begin{aligned}
\label{eq:q_belief_lip_constant}
\LQB &= \Rmax + \gamma (\Vmax + \LQB) \\
\Rightarrow  \quad   \LQB &= \frac{\Rmax + \gamma \Vmax}{1 - \gamma} \leq \frac{\Vmax}{1 - \gamma}
\end{aligned}
\end{equation}

We can now use the Lipschitz constant from ~\eref{eq:q_belief_lip_constant} in~\eref{eq:q_belief_temp}  :
\begin{equation}\label{eq:q_belief_bound}
\abs{ \Q(\stateaction, \belief_1) -  \Q(\stateaction, \belief_2)}  \leq \frac{\Vmax}{1 - \gamma} \norm{ \belief_1 - \belief_2 }{1}
\end{equation}

\vspace{1em}
\noindent\textbf{Lipchitz contiuous Q value}
\vspace{1em}

 We can now show that the Q value is Lipschitz continuous in state-belief-action space. For any two tuples $(\stateaction_1, \belief_1)$ and $(\stateaction_2, \belief_2)$ satisfying Assumption~\ref{assump:smooth_reward_transition} and Assumption~\ref{assump:smooth_reward_transition}, the following is true:

\begin{equation}
\begin{aligned}\label{eq:q_lipschitz_bound}
    & \abs{ \Q(\stateaction_1, \belief_1) - \Q(\stateaction_2, \belief_2) } \\
    & \leq | \rewardFn(\stateaction_1, \belief_1) - \rewardFn(\stateaction_2, \belief_2)\\
    & \quad + \gamma \sum\nolimits_{\state'} P(\state'|\stateaction_1,\belief_1) \V(\state',\belief_1') - P(\state' | \stateaction_2, \belief_2) \V(\state',\belief'_2)| \\
    &  \leq \LR d_{\stateaction_1,\stateaction_2} + R_{\max} \norm{ \belief_1 - \belief_2 }{1}\\
    & \quad + \gamma \Vmax \left( \LP  d_{\stateaction_1,\stateaction_2}  + \norm{ \belief_1 - \belief_2 }{1}\right)\\
    &\quad + \gamma  \sum\nolimits_{\state'} P(\state'|\stateaction_2,\belief_2) \max_{\action'} \abs{ \Q(\stateaction',\belief_1') - \Q(\stateaction',\belief'_2)}\\
    &\leq \left(\LR + \gamma \Vmax \LP \right) d_{\stateaction_1,\stateaction_2}  + (R_{\max} + \gamma \Vmax ) \norm{ \belief_1 - \belief_2 }{1} \\
    &\quad + \gamma \frac{\Vmax}{1 - \gamma} \norm{ \belief_1 - \belief_2 }{1} \\
    &\leq \left(\LR + \gamma \Vmax \LP \right) d_{\stateaction_1,\stateaction_2} \\
    &\quad + \left(R_{\max} + \frac{\gamma(2 - \gamma)}{1-\gamma} \Vmax  \right) \norm{ \belief_1 - \belief_2 }{1}
\end{aligned}
\end{equation}
where the 2nd inequality follows from taking similar steps as in \eqref{eq:q-continuous}, the 3rd inequality follows from \eqref{eq:q_belief_bound}  and Assumption~\ref{assump:belief_contraction}.

Now, with $\alpha > 0$, define the distance metric in state-belief-action space be the following:
\begin{equation}
\label{eq:belief_distance_metric}
d_{\stateaction_1,\belief_1,\stateaction_2,\belief_2} = \alpha d_{\stateaction_1,\stateaction_2} + \norm{ \belief_1 - \belief_2 }{1}
\end{equation}
From~\eref{eq:q_lipschitz_bound} and~\eref{eq:belief_distance_metric}, we can derive the following:
\begin{equation*}
  \LQ = \max \left(
    \frac{1}{\alpha} \left( \LR + \gamma \Vmax \LP \right),
    R_{\max} + \frac{\gamma(2 - \gamma)}{1-\gamma} \Vmax
  \right)
\end{equation*}


\subsection{Proof of \corref{corollary:escape}}

\begin{supplemma}[Lemma 56 in \cite{li2009unifying}]
\label{lemma: li-bernoulli} Let $x_1, ..., x_m \in \mathcal{X}$ be a sequence of $m$ independent Bernoulli trials, each with a success probability at least $\mu: E[x_i] \geq \mu$, for some constant $\mu > 0$ Then for any $l \in \mathcal{N}$ and $\delta \in (0,1)$, with probability at least $1-\delta$, $x_1 + ... + x_m \geq l$ if $m \geq \frac{2}{\mu}(l + \log\frac{1}{\delta})$.
\end{supplemma}

After at most $\frac{2\Qmax(1-\gamma)}{\epsilon}\left(k\cover(\epsD/\LtQ) + \log \frac{1}{\delta} \right)$ non-overlapping trajectories of length $T$, $\escapeEvent$ happens for at least $k\cover(\epsD/\LtQ)$ times with probability at least $1-\delta$. Then, from \lemref{lemma:kakade-known}, all reachable state-actions will have become known, making $P(\escapeEvent) = 0$. Setting $T = \frac{1}{1-\gamma} \log \frac{\Rmax}{\epsilon}$, we can have at most $m = \frac{2\Qmax}{\epsilon}\left(k\cover(\epsD/\LtQ) + \log \frac{1}{\delta} \right)\log \frac{\Rmax}{\epsilon}$ steps in which $P(\escapeEvent) \geq \frac{\epsilon}{\Qmax (1-\gamma)}$.


\subsection{Proof of \corref{corollary:near-optimality-known-BMDP}}

This follows from Lemma 3.13, 3.14 of \cite{pazis2013pac} to get $-\epsilon \leq \tQ - B\tQ \leq 2\epsilon$, and applying our \lemref{lemma:kakade-known}.

\subsection{Proof of \thmref{theorem:belief_cpace_pac}}

\begin{supplemma}[Lemma 2 in \cite{kearns2002near}]
\label{lemma: sufficient-T} If $T \geq \frac{1}{1-\gamma}\log\frac{\Rmax}{\epsilon}$, then $|V^\policy(\state,T) - V^\policy(\state)| \leq \frac{\epsilon}{1-\gamma}.$
\end{supplemma}

Our proof closely follows that of Theorem 3.16 of \cite{pazis2013pac}. Let $\knownBeliefMDP$ be the known belef MDP, and $\escapeEvent$ be the escape event. Let $\delta' = \frac{\delta}{2}$ and $k \in \left[\frac{\tQmax^2}{\epsilon^2} \log  \frac{2 \cover(\epsilon/\LtQ) }{ \delta' } , \frac{2 \cover(\epsilon/\LtQ)}{\delta'} \right]$, and let $T = \frac{1}{1-\gamma} \log \frac{\Rmax}{\epsilon}$.

At every step, either of the following events happen.

\begin{enumerate}
    \item $P(\escapeEvent) \geq \frac{\epsilon}{\Qmax (1-\gamma)}$:
From \corref{corollary:escape}, we can have at most $m = \frac{2\Qmax}{\epsilon}\left(k\cover(\epsilon/\LtQ) + \log \frac{1}{\delta'} \right)\log \frac{\Rmax}{\epsilon}$ steps in which $P(\escapeEvent) \geq \frac{\epsilon}{\Qmax (1-\gamma)}$.

    \item $P(\escapeEvent) < \frac{\epsilon}{\Qmax (1-\gamma)}$: With probability at least $1-\delta'$,
    \begin{align}
    \V^{\tpolicy_t}_{\beliefMDP}(\state_t, \belief_t) & \geq \V^{\tpolicy_t}_{\beliefMDP}(\state_t, \belief_t, T) \label{ineq: Rpositive}\\
    & \geq \V^{\tpolicy_t}_{\knownBeliefMDP}(\state_t, \belief_t, T) - \Qmax P(\escapeEvent) \label{ineq: EscapeMax}\\
    & \geq \V^{\tpolicy_t}_{\knownBeliefMDP}(\state_t, \belief_t, T) -  \frac{\epsilon}{1-\gamma} \label{ineq: Escape} \\
    & \geq  \V^{\tpolicy_t}_{\knownBeliefMDP}(\state_t, \belief_t) - \frac{2\epsilon}{1-\gamma} \label{ineq: Sufficient-T}\\
    & \geq V^*_{\knownBeliefMDP}(\state, \belief) - \frac{5\epsilon}{1-\gamma} \label{ineq: accuracy}\\
    & \geq  V^*_{\knownBeliefMDP}(\state, \belief, T) - \frac{5\epsilon}{1-\gamma} \label{ineq: Rpositive2}\\
    & \geq \V^{\tpolicy_t}_{\beliefMDP}(\state_t, \belief_t, T) -  \frac{6\epsilon}{1-\gamma} \label{ineq: Escape2}\\
    & \geq \V^{\tpolicy_t}_{\beliefMDP}(\state_t, \belief_t) -  \frac{7\epsilon}{1-\gamma} \label{ineq: Sufficient-T2}
    \end{align}
where \eqref{ineq: Rpositive}, \eqref{ineq: Rpositive2} come from rewards being nonnegative, \eqref{ineq: EscapeMax}, \eqref{ineq: Escape}, \eqref{ineq: Escape2} from \lemref{lemma:generalized-induced-inequality}, \eqref{ineq: Sufficient-T}, \eqref{ineq: Sufficient-T2} from \supplemref{lemma: sufficient-T}, and \eqref{ineq: accuracy} come from \corref{corollary:near-optimality-known-BMDP}. The proof of \thmref{theorem:belief_cpace_pac} comes from taking the union of these two cases.
\end{enumerate}


\subsection{Proof of Proposition~\ref{prop:seeding}}

Firstly, because we use exact seed values in  Definition~\ref{def:seeding_exact}, we do not need to add samples in the space covered by this estimate. Secondly, using this reduced number of samples does not violate any preconditions to the main theorem while leading to a lower sample complexity.

From Corollary~\ref{corollary:near-optimality-known-BMDP}, we need to show the following condition holds everywhere:
\begin{equation}
\label{eq:estimate_bellman_error}
	-\epsilon \leq \tQ(\stateaction,\belief) - B\tQ(\stateaction,\belief) \leq 2\epsilon
\end{equation}

We split the space of $(\stateaction, \belief)$ into two regions. Let region $\mathcal{R}_1$ be the space where we apply the exact seed estimate, i.e. $\mathcal{R}_1 = \setst{(\stateaction, \belief)}{\exists \onehot{i},  \dis{\belief}{\onehot{i}} \leq \frac{\epsilon}{\LQ(1 + \gamma)}}$. In this region, no samples are added. Let region $\mathcal{R}_2$ be the complement of $\mathcal{R}_1$ i.e. the reduced space referred to in Proposition~\ref{prop:seeding}. This is the only space where samples are added.

We first consider region $\mathcal{R}_1$. Here we use the estimate from Definition~\ref{def:seeding_exact}, i.e. $\tQ(\stateaction,\belief) = Q(\stateaction, \latent)$. From the Lipschitz property of $Q$, we have the following bound:
\begin{equation*}
\begin{aligned}
\label{eq:estimate_optimality_bound}
\norm{\tQ(\stateaction, \belief) - Q^*(\stateaction, \belief)}{\infty} & \leq \norm{Q(\stateaction, \latent) - Q^*(\stateaction, \belief)}{\infty} \\
& \leq \max_{\belief} \LQ \norm{\belief - \onehot{\phi}}{1} \\
& \leq \LQ \frac{\epsilon}{\LQ(1 + \gamma)} = \frac{\epsilon}{(1 + \gamma)}
\end{aligned}
\end{equation*}

We now use this to get bound the Bellman error:
\begin{equation*}
\begin{aligned}
& \norm{\tQ(\stateaction, \belief) - B \tQ(\stateaction, \belief)}{\infty} \\
& \leq \norm{\tQ(\stateaction, \belief) - Q^*(\stateaction, \belief)}{\infty} + \norm{B \tQ(\stateaction, \belief) - Q^*(\stateaction, \belief)}{\infty}\\ 
& \leq \norm{\tQ(\stateaction, \belief) - Q^*(\stateaction, \belief)}{\infty} + \gamma \norm{\tQ(\stateaction, \belief) - Q^*(\stateaction, \belief)}{\infty} \\
& \leq (1 + \gamma) \norm{\tQ(\stateaction, \belief) - Q^*(\stateaction, \belief)}{\infty} \\
& \leq (1 + \gamma) \frac{\epsilon}{1 + \gamma} = \epsilon 
\end{aligned}
\end{equation*}
where the first inequality comes from the triangle inequality, the second one comes from the fact that $B$ is a $\gamma$ contraction, and the last inequality comes from \eref{eq:estimate_optimality_bound}. Hence the Bellman error of the estimate~\eref{eq:estimate_bellman_error} holds in $\mathcal{R}_1$.

Now, let $\coverReduced(\epsD/\LtQ)$ be the covering number of  $\mathcal{R}_2$. With
$\sample = \frac{2 \Qmax}{\epsilon}  \left( k \coverReduced(\epsilon/\LtQ) + \log \frac{2}{\delta}  \right)\log \left( \frac{\Rmax}{\epsilon} \right)$
samples, the Bellman error of the estimate~\eref{eq:estimate_bellman_error} holds in $\mathcal{R}_2$.


\subsection{Proof of \propref{prop:approx_vi}}

The procedure described in Definition~\ref{def:optimistic_estimate}, which we call an optimistic Bellman operator $\tilde{B}$, is a contraction. Let $\tQ$ be an estimate, $\tB^{\infty} \tQ$ be the fixed point solution, and $\norm{\tQ - \tB^{\infty} \tQ}{\infty} \leq \epsilon'$. Then from Lemma 3.15 in~\cite{pazis2013pac}, we have:
\begin{equation*}
 	\norm{\tB\tQ - \tB^{\infty} \tQ}{\infty} \leq \gamma \epsilon'
\end{equation*}

Note that $\norm{ \tQ - \tB^\infty \tQ }{\infty} \leq \tQmax$. Hence to reach $\norm{ \tB^i \tQ - \tB^\infty \tQ }{\infty} \leq \tolVI$, we need $i$ iterations such that
\begin{equation*}
	\gamma^i \tQmax \leq \tolVI ,\quad i \geq \frac{ \log \left( \nicefrac{\tolVI}{\tQmax} \right) }{\log \gamma}
\end{equation*}

We now examine the consequences of this gap. From pre-conditions in Corollary~\ref{corollary:near-optimality-known-BMDP}, we have
\begin{equation*}
\begin{aligned}
	-\epsilon &\leq \quad \tB^\infty \tQ - B\tQ &&\leq 2\epsilon \\
	-\epsilon - \tolVI &\leq \quad \tB^i \tQ - B\tQ &&\leq 2\epsilon + \tolVI\\
\end{aligned}
\end{equation*}

An updated Corollary~\ref{corollary:near-optimality-known-BMDP} with this inflated gap, results in an updated suboptimality gap in the main theorem:
\begin{equation*}
\V^{\tpolicy_t}(\state_t, \belief_t) \geq \V^*(\state_t, \belief_t) - \frac{7\epsilon + 2 \tolVI}{1 - \gamma}
\end{equation*}

\subsection{Proof of \propref{prop:batch-update}}

The suggested change in the algorithm implies that the known belief MDP $\knownBeliefMDP$ upon which the policy is built is fixed for every $nT$ steps. Let $\knownSet_1, \knownSet_2, \cdots$ be the known sets at every $nT$ steps. Suppose we only consider the first length $T$ trajectory of each $nT$ intervals when evaluating whether an escape happened. Suppose $P(\escapeEvent) >  \frac{\epsilon}{\Qmax(1-\gamma)}$ for some subset of $\knownSet_1, \knownSet_2, \cdots $. From \supplemref{lemma: li-bernoulli}, after at most $\frac{2\Qmax(1-\gamma)}{\epsilon}\left(k\cover(\epsD/\LtQ) + \log \frac{1}{\delta} \right)$ such $\knownSet$s, $\escapeEvent$ happens for at least $k\cover(\epsD/\LtQ)$ times with probability at least $1-\delta$. Setting $T = \frac{1}{1-\gamma} \log \frac{\Rmax}{\epsilon}$ as in \corref{corollary:escape}, we can have at most $m = \frac{2n\Qmax}{\epsilon}\left(k\cover(\epsD/\LtQ) + \log \frac{1}{\delta} \right)\log \frac{\Rmax}{\epsilon}$ steps in which $P(\escapeEvent) \geq \frac{\epsilon}{\Qmax (1-\gamma)}$.

\clearpage

\end{document}